\def\year{2019}\relax
\documentclass[letterpaper]{article} 

\usepackage{aaai19}  
\usepackage{times}  
\usepackage{helvet}  
\usepackage{courier}  
\usepackage{url}  
\usepackage{graphicx}  
\usepackage{amsfonts}
\usepackage{amssymb,amsmath,array}
\usepackage{multirow}
\usepackage{xcolor}
\begin{document}
%
\title{Exploiting Synthetically Generated Data with Semi-Supervised\\Learning for Small and Imbalanced Datasets}
\author{M. P\'erez-Ortiz\\
Department of Computer Science\\University College London\\66-72 Gower Street Bloomsbury\\ London (UK) CB3 0FD\\ maria.perez@ucl.ac.uk
\And P. Ti\v{n}o\\
School of Computer Science\\University of Birmingham\\Edgbaston\\ Birmingham (UK) B15 2TT\\ p.tino@cs.bham.ac.uk
 \And R. Mantiuk\\
Department of Computer\\Science and Technology,\\University of Cambridge\\15 JJ Thomson Avenue\\ Cambridge (UK) CB3 0FD\\ rkm38@cam.ac.uk
\And C. Herv\'as-Mart\'inez\\
Department of Computer Science\\and Numerical Analysis,\\University of C\'ordoba (Spain)\\Rabanales Campus\\C2 building 14071\\ chervas@uco.es
}
\maketitle
%
%

\begin{abstract}
Data augmentation is rapidly gaining attention in machine learning.
Synthetic data can be generated by simple transformations or through the data distribution.
In the latter case, the main challenge is to estimate the label associated to new synthetic patterns.
This paper studies the effect of generating synthetic data by convex combination of patterns and the use of these as unsupervised information in a semi-supervised learning framework with support vector machines, avoiding thus the need to label synthetic examples. We perform experiments on a total of 53 binary classification datasets. Our results show that this type of data over-sampling supports the well-known cluster assumption in semi-supervised learning, showing outstanding results for small high-dimensional datasets and imbalanced learning problems. 
\end{abstract}

\section{Introduction}

\frenchspacing


One of the current challenges in machine learning is the lack of sufficient data \cite{Forman2004}. In this scenario, over-fitting becomes hard to avoid, outliers and noise represent an important issue and the model generally has high variance. Several approaches have been proposed to deal
with small datasets, although the work in this matter is still scarce. From all the proposed approaches, synthetic sample generation or data augmentation techniques \cite{Li:2014:GAV:2658282.2658333,DBLP:journals/corr/WongGSM16,Yang2011740} have shown competitive performance, acting as a regulariser \cite{DBLP:journals/corr/abs-1710-09412}, preventing over-fiting and improving the robustness of both classifiers and regressors.


The generation of virtual examples is highly nontrivial and has been studied from different perspectives. Proposed methods use prior information \cite{Niyogi98}, add noise \cite{DBLP:journals/corr/abs-1710-09412}, apply simple transformations \cite{Ciresan:2010:DBS:1943016.1943024,Simard:2003:BPC:938980.939477,NIPS2012_4824,7298594} or use data over-sampling approaches \cite{Chawla02smote:synthetic,Perez2016}.

The most straightforward over-sampling approach is to randomly replicate data. However, this can lead to over-fitting \cite{Galar2012}. Another common approach is to do over-sampling taking into account the data distribution. A convex combination of patterns close in the input space has been successfully used for that purpose \cite{Chawla02smote:synthetic,DBLP:journals/corr/abs-1710-09412,Perez2016}. 


In this paper
we investigate the benefits and limitations of this simple
data augmentation technique coupled with SSL support vector machines. The motivations for such an approach are: i) when performing over-sampling one of the biggest challenges is how to label synthetic examples (potentially alleviated when using SSL as no label is assumed) and ii) the hypothesis that over-sampling by convex combination of patterns can support the cluster assumption in SSL and help to simplify the classification task.
The cluster assumption states that high density regions with different class labels must be separated by a low density region. Given this, two patterns are likely to have the same class label if they can be connected by a path passing through high density regions. 
The method proposed here is based on the synthetic generation of high density regions as an inductive bias for the classifier. 
We perform a thorough set of experiments over $27$ synthetic and $26$ benchmark binary datasets, showing
how this approach helps to mitigate the effect of small, high-dimensional and imbalanced datasets.



%

\section{Methodology}


\subsection{Data over-sampling by convex combination}

Assume that data forms a finite sample $X = \{\mathbf{x}_1,\ldots,\mathbf{x}_n\} \sim$ i.i.d. from a distribution $F$ and that our aim is to construct a finite-sample function of $X$. Resampling approximates the finite-sample distribution of the function computed over $X$ by the exact distribution of the function over $X^*$:
\begin{equation}\small
 X^*=\{\mathbf{x}_1^*,\ldots,\mathbf{x}_m^*\} \; \; \sim F^*(\mathbf{x}_1,\ldots,\mathbf{x}_n),
\end{equation}
where $F^*$ is defined as the resampling distribution and explicitly depends on the observations in $X$. 
Resampling is commonly used in machine learning for data augmentation. 


%

In the case of binary classification we also have access to a labelling $Y = (y_1, \ldots, y_n) \; \in \{-1,1\}^n$. When dealing with small or imbalanced datasets, appropriately capturing the joint probability function $P(X,Y)$ might be unrealistic. Because of this, most over-sampling approaches are rather simple. Usually, synthetic patterns are generated by convex combination of two seed patterns belonging to the same class and labelled directly using the same class label \cite{Chawla02smote:synthetic}. The first seed pattern $\mathbf{x}_i$ is chosen randomly, and the second one is chosen as one of its $k$-nearest neighbours. 
 $k$ is responsible for avoiding label inconsistencies and exploiting the local information of the data, but it can also significantly limit the diversity of synthetic patterns. 

\subsubsection{Limitations}
\begin{figure}[ht!]
   \centering
     \includegraphics[width = 0.37\textwidth]{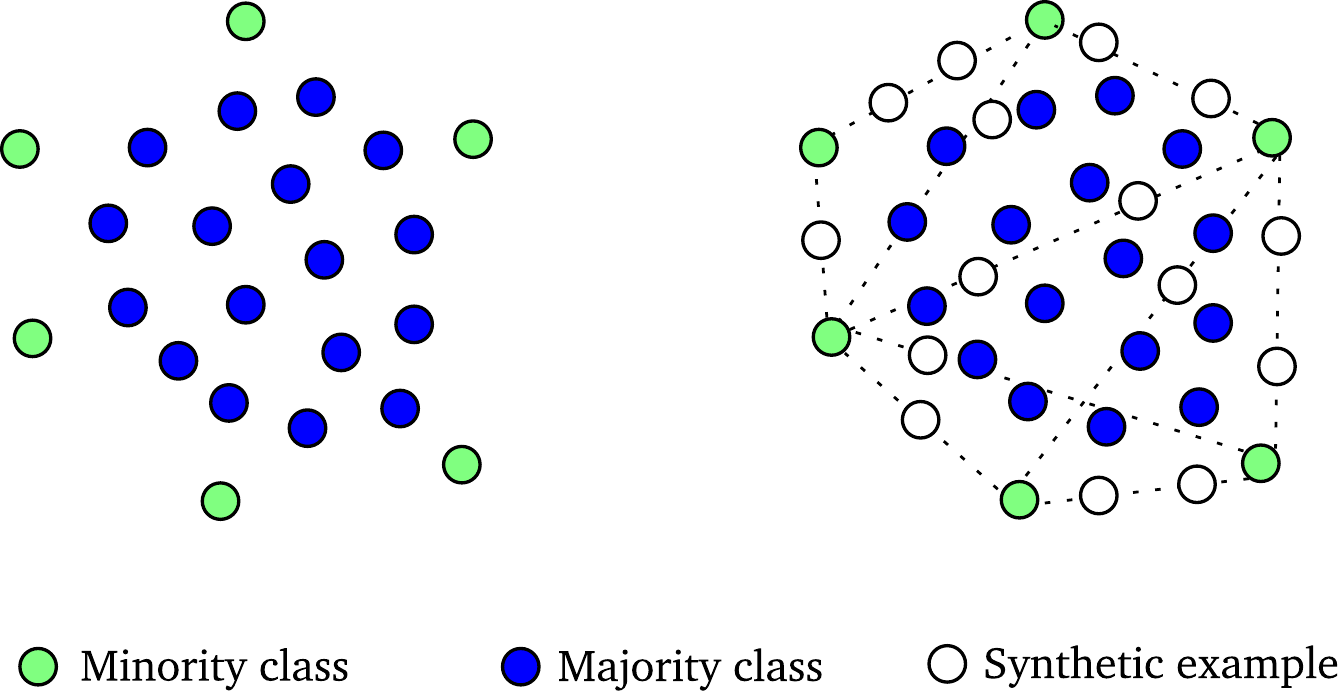}
  \caption{Example of an over-sampled imbalanced dataset, in which naively labelling synthetic data as minority class patterns might not be suitable.}
\label{fig:problem}
 \end{figure}

 \figurename{ \ref{fig:problem}} shows a toy imbalanced dataset where the classes are not convex (left) and some examples of synthetic data patterns that could be created for the minority class in order to balance the class distributions (right). This shows
 a representation of the main problem encountered when using this over-sampling approach, especially when the parameter of $k$-nearest neighbour is not properly optimised: synthetic patterns are created in the region of the majority class and if we naively label these patterns as minority class patterns, we introduce what we denote as a label inconsistency.

Different reformulations of SMOTE have been proposed to solve this. One of the recent proposals to do so is to perform synthetic over-sampling in
the feature space induced by a kernel function \cite{Perez2016}. In this new feature space the classes would be ideally linearly separable (and convex) and label inconsistencies can be avoided when using a convex combination of patterns. This technique has been seen to significantly improve the standard SMOTE.

\subsubsection{Effect on the data distribution}\label{sec:effect}

We study now the effect of over-sampling by means of a convex combination of patterns. At every step $j= 1,\ldots,m$ we create a synthetic instance $\mathbf{x}^*_j$ by selecting at random two patterns $\mathbf{x}_i,\mathbf{x}_h $:
\begin{eqnarray}\label{eq:convexcomb} \small
&\mathbf{x}^*_j = \mathbf{x}_i + (\mathbf{{x}}_h - \mathbf{x}_i)\cdot \delta_j =\\&= \delta_j \mathbf{x}_h + (1-\delta_j)\mathbf{x}_i, \;\; \delta_j \in U[0,1], \;\; \mathbf{x}^*_j \sim F^* \nonumber,
\end{eqnarray}
we restrict $\mathbf{{x}}_h \in k$-$nn(\mathbf{{x}}_i)$, where $k$-$nn$ represents a function that returns the $k$-nearest neighbours of $\mathbf{x}_i$.  Note that when over-sampling within a classification framework $\mathbf{x}_h$ is usually also restricted so that $y_h = y_i$. 


For simplicity, let us first assume $X \subseteq  \mathbb{R}$ and ${x}_i$ and ${x}_h$ come from the same Normal distribution $x_i,x_h \sim \mathcal{N}( \mu,\sigma^2)$. The definition of the characteristic function of the Normal distribution is: 
\begin{equation}\small
 \varphi_X(it) = E[e^{itX}] = e^{i\mu t - \frac{\sigma^2 t^2}{2}}.
\end{equation}

The new random variable ${x}^* = \delta_j {x}_h + (1-\delta_j){x}_i$ will have the characteristic function: 
\begin{eqnarray}\small
\nonumber
 &\varphi_{\delta_j {x}_h + (1-\delta_j){x}_i}(it) = E[e^{it(\delta_j {x}_h + (1-\delta_j){x}_i)}] =\\ \nonumber
 &= E(e^{it\delta_j {x}_h})E(e^{it(1-\delta_j){x}_i})  =\\ \nonumber
  &= e^{i\mu \delta_j t - \frac{\sigma^2 \delta_j^2 t^2}{2}}e^{i\mu(1-\delta_j)t - \frac{\sigma^2 (1 - \delta_j)^2 t^2}{2}} =\\&=   e^{i\mu t - \frac{\sigma^2 (1 - 2\delta_j + 2\delta_j^2) t^2}{2}},
\end{eqnarray}
meaning that the convex combination of these two patterns will follow the distribution: $x_j^* \sim \mathcal{N}( \mu,\sigma^2\cdot (1-2\delta_j+2\delta_j^2) )$,
which for $\delta_j \sim U[0,1]$ translates into $(1-2\delta_j+2\delta_j^2)$ being within $[0.5,1]$. This means that the resampled distribution $F^*$ will most probably have a lower variance, yielding synthetic data more concentrated around the mean. 


If seed patterns do not come from the same distribution, i.e. $x_i \sim \mathcal{N}( \mu_i,\sigma_i^2)$ and $x_h \sim \mathcal{N}( \mu_j,\sigma_j^2)$, then $x_j^* \sim \mathcal{N}(\delta_j \mu_h + (1-\delta_j)\mu_i,\delta_j^2 \sigma_h^2 + (1-\delta_j)^2\sigma_i^2)$. We assume, however, that given that these patterns are neighbours, they do come from the same distribution.

The density function of $X^*$ assuming $\delta \sim U[0,1]$ is:
\begin{eqnarray}\small
\nonumber
&\tilde{p}({x}^*) = \int_{0}^{1} p(\delta) \cdot f({x}^*|\mu,\sigma^2(1-2\delta + 2\delta^2)) d\delta =\\ \nonumber  &=\int_{0}^{1} f({x}^*|\mu,\sigma^2(1-2\delta + 2\delta^2)) d\delta=\\  &= \frac{1}{\sqrt{2\pi}} \int_{0}^{1} \frac{1}{\sqrt{\sigma^2(1-2\delta + 2\delta^2)}} \cdot e^{\left ( -\frac{({x}^*-\mu)^2}{2\sigma^2(1-2\delta + 2\delta^2)}\right )} d\delta,
\end{eqnarray}
$f$ being the density function of the Normal distribution and the density function $p(\delta)=1$.
The variance of $X^*$ can thus be evaluated as:
\begin{eqnarray}\small
 &V[X^*] = \int_{-\infty}^\infty ({x^*}-\mu)^2 \cdot \tilde{p}({x^*}) \cdot d{x}^*  =\\ &= \int_{-\infty}^{+ \infty} \int_0^1 ({x^*}-\mu)^2  f({x}^*|\mu,\sigma^2(1-2\delta + 2\delta^2))  d\delta  d{x}^* \nonumber
 \end{eqnarray}
This integral can be numerically evaluated. When doing so we see that the original variance is always reduced by 0.333. 

Given that over-sampling is applied independently per dimension, we have: 
$
 \tilde{p}(\mathbf{x}^*) = \prod_{i=1}^d \tilde{p}_i({x}_{(i)}^*),
$
where ${x}_{(i)}$ is the i-th dimension of $\mathbf{x}$. 

Let us now analyse the multivariate case where $X \subseteq \mathbb{R}^d$, $d>1$ and $\mathbf{x} \sim \mathcal{N}({\boldsymbol{ \mu},\boldsymbol {\Sigma_{\mathbf{x}} }})$. For that let us first assume a matrix $\mathbf{P}$ for changing the basis such that $\mathbf{z} = \mathbf{P} \mathbf{x}$. If we choose $\mathbf{P}$ to be a basis formed by the unit eigenvectors of $\Sigma_{\mathbf{x}}$ then it is easy to show that $\Sigma_{\mathbf{z}}$ (i.e. the covariance matrix of $\mathbf{z}$) is a diagonal matrix formed by the eigenvalues associated to $\Sigma_{\mathbf{x}}$, i.e. the i-th diagonal value $\lambda_i$ is the variance of $\mathbf{x}$ along the i-the eigenvector $\mathbf{p}_i$ of $\mathbf{P}$. In the rotated axis Eq. \ref{eq:convexcomb} can be rewritten as:
\begin{equation}\small
\mathbf{z}_j^* \equiv \mathbf{P} \mathbf{x}^*_j = \delta_j \mathbf{P} \mathbf{x}_h + (1-\delta_j) \mathbf{P}\mathbf{x}_i,
\end{equation}
since $P$ is a linear operator. Convex combinations of patterns are thus invariant to rotations of the co-ordinate axis.
In this axis, the data coming from our transformed resampling distribution $\mathbf{z}^* \sim \mathcal{N}(\boldsymbol{ \mu},\boldsymbol {\tilde{\Sigma}_{\mathbf{z}^*} })$ will have the diagonal covariance matrix:
 \begin{equation}\small
\boldsymbol {\tilde{\Sigma}_{\mathbf{z}^*} } =
 \begin{pmatrix}
 \lambda_1 (1 - 2\delta_j + 2\delta_j^2) & \ldots & 0\\ 
0  & \ldots & 0 \\ 
0  & \ldots & \lambda_d (1 - 2\delta_j + 2\delta_j^2)
\end{pmatrix}
 \end{equation}
 It follows that when over-sampling through convex combinations of patterns using the uniform distribution the mean of the data will remain unchanged and so will the eigenvectors of the covariance matrix, but the eigenvalues will shrink.

\begin{figure}[ht!]
   \centering
     \includegraphics[width = 0.35\textwidth]{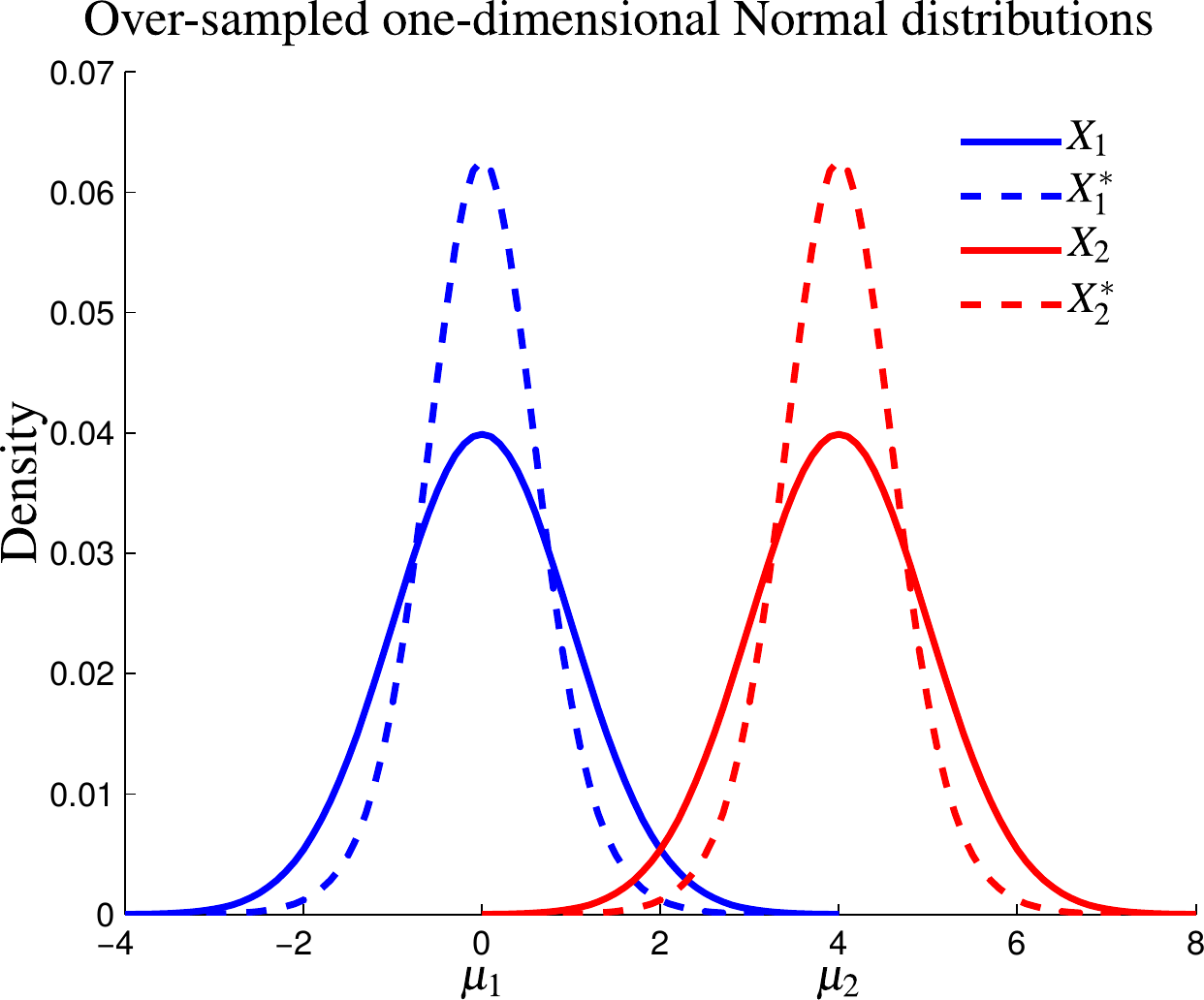}
  \caption{Normal distributions and class dependent over-sampled distributions (dotted line).}
\label{fig:twoclassnormals}
 \end{figure}

 \figurename{ \ref{fig:twoclassnormals}} shows the result of over-sampling two Normal distributions, where $X_i$ represents the data associated to class $\mathcal{C}_i$ in our classification problem. It can be seen that by performing convex combinations of patterns we change the data distribution. We use this to induce high-density regions that are later used by the SSL algorithm. 

\subsection{Semi-supervised learning (SSL)}

In semi-supervised learning (SSL), we not only have access to $n$ labelled data $\mathcal{L} = (\mathbf{x}_i, y_i)_{i=1}^n$ drawn i.i.d. according to $P(X,Y)$, but also to $m$ additional unlabelled data $\mathcal{U} = \{\mathbf{x}^u_i\}_{i=1}^m$ drawn i.i.d. according to $P(X)$. 


Up to this date, theoretical analysis of SSL fails to provide solid evidence for the benefits of using unlabelled patterns in a supervised learning task \cite{Ben-david08d.:does}. Generally, the consensus reached in the literature is that unlabelled data: (i) should be used with care because it has been seen to degrade classifier performance in some cases (e.g. when we assume incorrect data models \cite{Cozman:2003:SLM:3041838.3041851} or there are outliers or samples of unknown classes \cite{312897}; (ii)
 is mostly beneficial in the presence of a few labelled samples \cite{conf/nips/SinghNZ08,312897,Cozman:2003:SLM:3041838.3041851};
 (iii) can help
to mitigate the effect of the Hughes phenomenon (i.e. the curse of dimensionality) \cite{312897}; 
(iv) can help only if there exists a link between the marginal data distribution and the target function to be learnt and both labelled and unlabelled data are generated from the same data distribution \cite{Huang:2006:CSS:2976456.2976532}; and finally (v) can improve on the performance of supervised learning when density sets are discernable from unlabelled but not from
labelled data \cite{conf/nips/SinghNZ08}. 


SSL algorithms can be classified using the following taxonomy \cite{Chapelle:2010:SL:1841234}: i) Generative models which estimate the conditional density $P(X|Y)$; ii) low density separators that maximise the class margin; iii) graph-based models which propagate information through a graph; and finally, iv) algorithms based on a change of representation. The most widely used SSL algorithms belong to the low density separators or the graph-based models groups. Generative approaches are said to solve a more complex problem than discriminative ones and require more data and the algorithms based on a change of representation do not use all the potential of unlabelled data. Because of this, we focus on low density separators.

  \begin{figure}[ht!]
   \centering
     \includegraphics[width = 0.45\textwidth]{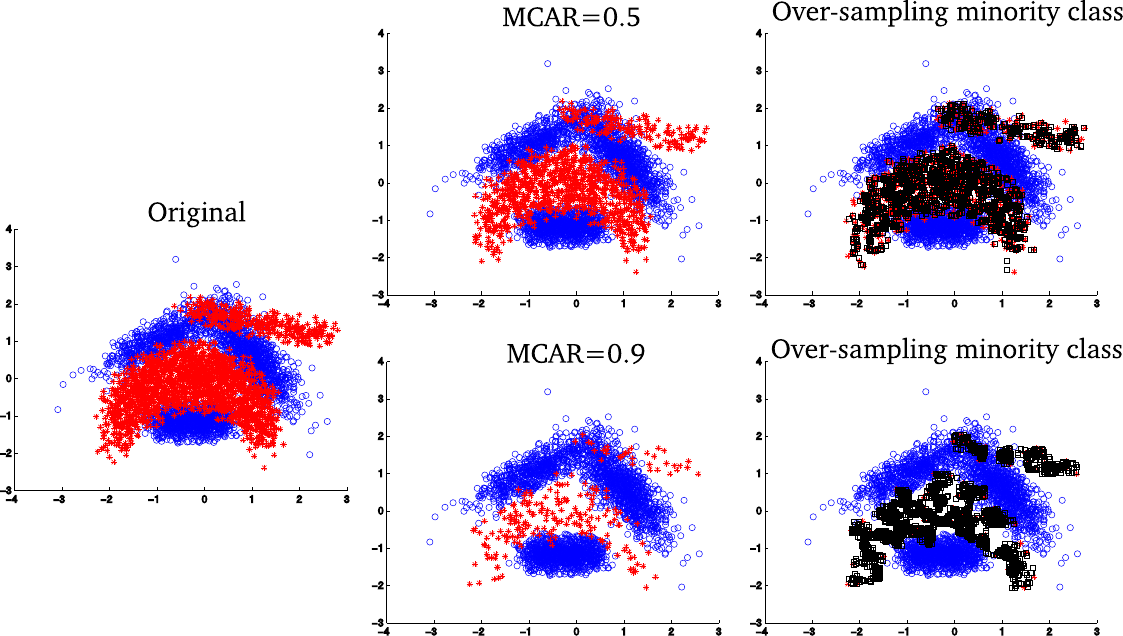}
  \caption{Over-sampling in the banana dataset. Left figure shows the original dataset, where colours indicate the class. The figures in the middle show the dataset where data is missing at random (MCAR) for one of the classes with percentages of missing patterns of 50\% and 90\%. The figures on the right show the over-sampled datasets. }
\label{fig:banana}
 \end{figure}

\subsection{Exploiting the cluster assumption}

 Labelling synthetically generated patterns without knowledge about $P(X,Y)$ is a highly nontrivial problem. Instead, we approach this by using SSL, assuming that every synthetic pattern belongs to the set of unlabelled data, $\mathbf{x}_j^*\in \mathcal{U}$.

We exploit the cluster assumption by artificially connecting labelled patterns $\mathbf{x}_i$ and $\mathbf{x}_h $ belonging to the same class ($y_i = y_j$) through unlabelled samples. 
Two patterns $\mathbf{x}_i$ and $\mathbf{x}_h$ are said to be connected if there exist a sequence of
relatively dense patterns such that the marginal density $P(X)$ varies smoothly along the sequence of patterns between $\mathbf{x}_i$ and $\mathbf{x}_h$ \cite{conf/nips/SinghNZ08}.
We have shown in Section \ref{sec:effect} that over-sampling two patterns $\mathbf{x}_h$ and $\mathbf{x}_i$ by convex combination makes the density function more compact in the region that connects them. This property is maintained for all random variables that are a linear combination of two patterns $\mathbf{x}_h$ and $\mathbf{x}_i$ that come from the same distribution (independently on whether their distribution has the reproductive property). 
The cluster assumption is the basis for different low-density semi-supervised learners. This assumption implies that if two patterns are linked by
a path of high density (e.g., if they belong to the same cluster), then their outputs
are likely to be similar \cite{Chapelle:2010:SL:1841234}. Our proposal of using $X^*$ as unlabelled samples can thus be seen as synthetically generating high density regions for each class as
an inductive bias for the classifier. An example of over-sampling can be found in \figurename{ \ref{fig:banana}} where over-sampled patterns are plotted in black.

%
  Our objective is thus to seek a classifier $g$ and a labelling $\mathbf{y}^*=\{y^*_1,\ldots,y^*_m\} \in \{-1,+1\}^m$ by minimising: 
\begin{eqnarray}\small
 \arg \min_{g,\mathbf{y}*} \frac{\lambda}{n} \sum_{i=1}^n {L}(y_i \cdot g(\mathbf{x}_i)) + \frac{\lambda^*}{m} \sum_{j=1}^m {L}^*(y^*_j \cdot g(\mathbf{x}^*_j)). \label{eq:lds}
\end{eqnarray}
where ${L},{L}^*:\mathbb{R} \rightarrow \mathbb{R}$ are loss functions and $\lambda$ and $\lambda^*$ are real-valued parameters which reflect confidence in labels and the cluster assumption respectively. The labels of synthetic data are treated as additional optimisation variables, as it is common in SSL \cite{sindhwani2006deterministic,sindhwani2006large}.
An effective loss function ${L}^*$ over an unlabelled pattern $\mathbf{x}^*_j$ is ${L}^*(g(\mathbf{x}^*_j)) = \min \{{L}(g(\mathbf{x}^*_j)),{L}(-g(\mathbf{x}^*_j))\}$, which corresponds to making the optimal choice for unknown label $y^*_j$ and promotes decision boundaries that pass through low-density regions.

\subsubsection{Choice of low density separator}

The most common approach for constructing a SSL low density separator is to use a maximum margin
approach (e.g. using Support Vector Machines, SVMs). 
However, the formulation in Eq. \ref{eq:lds} results in a hard optimisation problem when unlabelled data is abundant.
In the semi-supervised SVM classification setting (S$^3$VM), this minimisation problem is solved over both the hyperplane parameters $(\mathbf{w},b)$ and the label vector $\mathbf{y}^*$, 
\begin{equation}\label{s3vm} \small
  \arg \min_{(\mathbf{w,b}),\mathbf{y}^*} \frac{1}{2} ||\mathbf{w}||^2 +  \lambda \sum_{i=1}^n V(y_i,o_i) + \lambda^*  \sum_{j=1}^m V(y^*_i,o^*_j),
\end{equation}
where $o_i = \mathbf{w}^T\mathbf{x}_i + b$ and V is a loss function. This problem is solved under the class balancing constraint: 
\begin{equation} \small
 \frac{1}{m} \sum_{i=1}^m \max(y^*_i,0) = r,
\end{equation}
where $r$ is a user-specified ratio of unlabelled data to be assigned to the positive class. Unlike SVMs, this S$^3$VM formulation leads to a non-convex optimization problem, which is solved either by combinatorial or continuous optimisation \cite{Chapelle:2008:OTS:1390681.1390688}. 

The method chosen in this paper is S$^3$VM$^\texttt{light}$, which has shown promising performance and is robust to changes in the hyperparameters \cite{Chapelle:2008:OTS:1390681.1390688}. This technique is based on a local combinatorial search guided by a label switching procedure. The vector $\mathbf{y}^*$ is initialised as the labelling given by a SVM trained only on the labelled set. This labelling is restricted to maintain the class ratios previously defined by $r$. Subsequent steps of the algorithm comprise of switching the labels of two unlabelled patterns $\mathbf{x}^*_j$ and $\mathbf{x}^*_z$ (in order to maintain class proportions) that satisfy the following condition: 
\begin{eqnarray}\nonumber \small
 &y^*_j = 1, y^*_z = -1 \\ \small &V(1,o^*_j) + V(-1,o^*_z) > V(-1,o^*_j) + V(1,o^*_z),
\end{eqnarray}
i.e. the loss after switching these labels is lower. 

Concerning the computational complexity of our proposal, the main bottleneck is the SSL part as the complexity of
over-sampling is linear. The complexity of S$^3$VM$^\texttt{light}$ is of the same order as that of a standard SVM.
However, it will be trained with more data (i.e. real plus synthetic).

\subsubsection{Ensemble of synthetic hypotheses}
Since
the estimation of the resampling distribution $F^*$ is a stochastic process, we also consider the use of different resampling distributions in an ensemble framework. 
The application is straightforward: each member of the ensemble is formed by a resampling distribution $F^*$ and a S$^3$VM model $(\mathbf{w},b)$. Final labels are computed by majority voting. 

\begin{table}[t!]
\centering
\fontsize{9.0pt}{10.0pt}
\selectfont
\caption{Characteristics for the $26$ benchmark datasets.}
\begin{tabular}{l c c l c c } \hline \hline
Dataset & $N$ & $d$ &  Dataset & $N$ & $d$ \\ \hline
haberman (HA)	&	306	&	3	&	hepatitis (HE)	&	155	&	19 	\\
listeria (LI)	&	539	&	4 	&	bands (BA)	&	365	&	19 	\\
mammog. (MA)	&	830	&	5 	&	heart-c (HC)	&	302	&	22 	\\
monk-2 (MO)	&	432	&	6	 &	labor	(LA) &	57	&	29 	\\
appendicitis (AP)	&	106	&	7 	& pima (PI)	&	768	&	8 	\\
glassG2	 (GL) &	163	&	9	&	credit-a (CR)	&	690	&	43 	\\
saheart	(SA) &	462	&	9	 &	specfth. (SP)	&	267	&	44 	\\
breast-w (BW)	&	699	&	9 		&	card (CA)	&	690	&	51 	\\
heartY	(HY) &	270	&	13 	&	sonar	(SO) &	156	&	60 	\\
breast	(BR) &	286	&	15	 &	colic (CO)	&	368	&	60	\\
housevot. (HO)	&	232	&	16  &	credit-g (CG)	&	1000	&	61\\ 
banana & 5300 & 2 & ionosphere & 351 & 34 \\ 
liver & 583 & 10 & wisconsin & 569 & 32 \\ 
 \hline \hline
\multicolumn{6}{l}{All nominal variables are transformed into binary ones}
\end{tabular}
\label{table:datasets}
\end{table}

 \section{Experimental results}
 
 In our experiments we try to answer the following questions: 
\begin{enumerate}
 \item What are the largest contributing factors to the degradation in performance when dealing with small datasets?
\item Does over-sampling prevent the need for collecting further data in small and imbalanced scenarios? 
\item How does our approach of using SSL and not labelling data compares to other approaches in the literature? 
\item In the context of classification, is it class dependent over-sampling better than class-independent? 
\end{enumerate}

To answer the first question, we do a first experiment using $27$ synthetically generated datasets. 
To answer questions 2-4, we perform two additional experiments, in which we test a wide range of approaches with $26$ real-world benchmark datasets, changing the percentage of missing patterns to study the influence of the data sample size (second experiment) and imbalanced class distributions (third experiment).

 All the methodologies have been tested considering the paradigm of Support Vector Machines (SVM) \cite{Cortes1995}. The $26$ benchmark datasets are extracted from the {UCI repository \cite{Lichman:2013}} (characteristics shown in \tablename{ \ref{table:datasets}}).
These datasets are not originally imbalanced or extremely small. Instead, these characteristics are generated synthetically by removing a percentage of patterns at random, so that the performance can be compared against the one with the original full dataset. 


Because of space restrictions, we only show mean test results and rankings, but all results can be accessed online\footnote{\url{https://doi.org/10.17863/CAM.32312}}.

 \subsection{Methodologies tested}
In order to address the difference between using real vs. synthetic data, 
we compare standard supervised SVMs (with no over-sampling or data missing) to different approaches with data Missing Completely At Random (MCAR).
Note that this comparison is not strictly fair, but it provides a useful baseline performance to evaluate our over-sampling approaches. Thus, our objective is not to surpass the performance achieved with real data by the use of synthetic one, but rather to reach a similar performance.
We also compare our proposed approach to: 1) previous over-sampling approaches that use naive labelling \cite{Chawla02smote:synthetic,Perez2016} and 2) transductive graph-based SSL, as another alternative for labelling synthetic data. Within our proposed methods we have different approaches: class-dependent and independent over-sampling (i.e. over-sampling classes separately or not) and an ensemble of $51$ S$^3$VM models using unlabelled synthetically generated patterns. Note that the optimisation procedure of SVM and S$^3$VM is different, which may influence the results (S$^3$VM is said to be more prone to reach local optima). Because of this, we include another approach as a baseline: S$^3$VM model that reintroduces the real data removed at random in the unsupervised set. The main purpose here is to compare over-sampled vs. real data within the S$^3$VM framework.

   \begin{figure}[ht!]
   \centering
     \includegraphics[width = 0.45\textwidth]{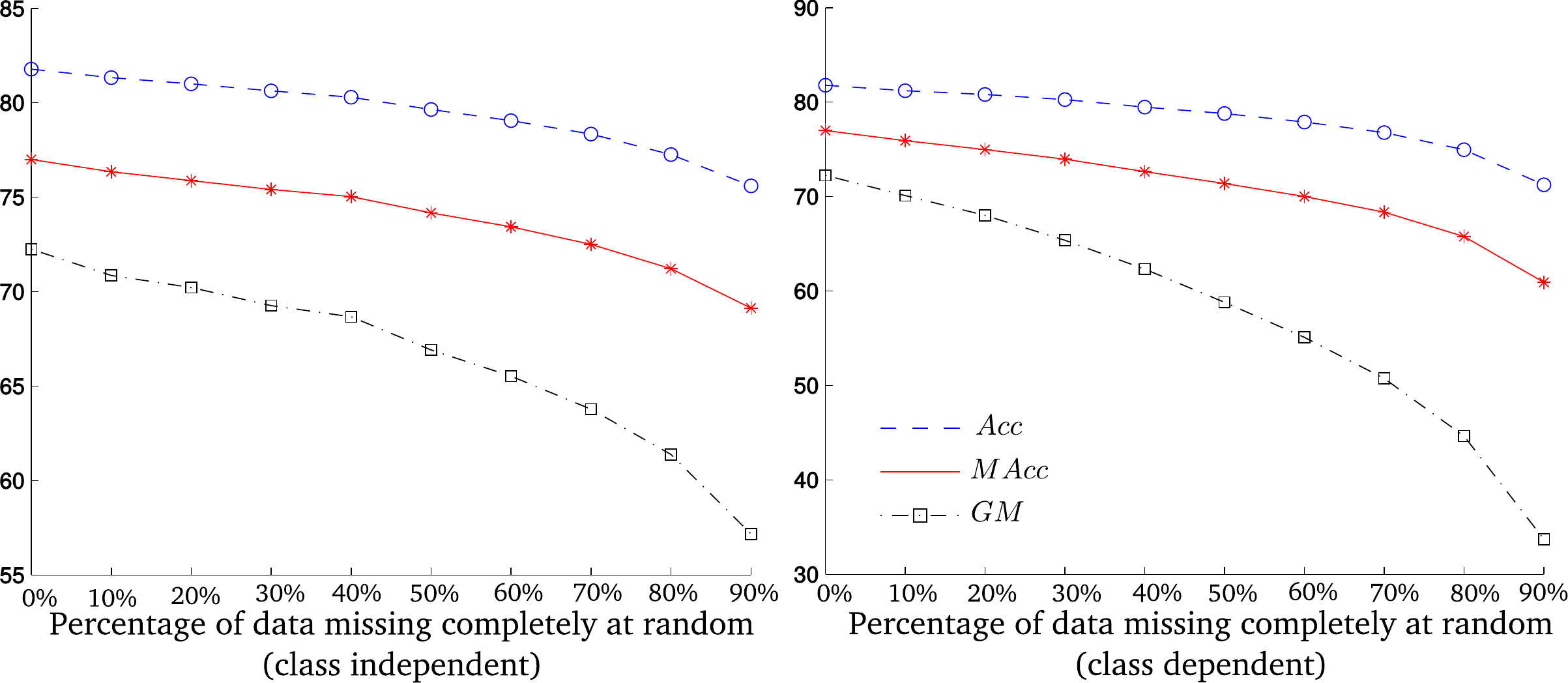}
  \caption{Mean test performance across all benchmark datasets for S-MCAR. In the left plot patterns are removed from both classes, whereas in the right plot patterns are removed only for the minority class.}
\label{fig:mcar}
 \end{figure}
  \begin{figure}[ht!]
   \centering
     \includegraphics[width = 0.45\textwidth]{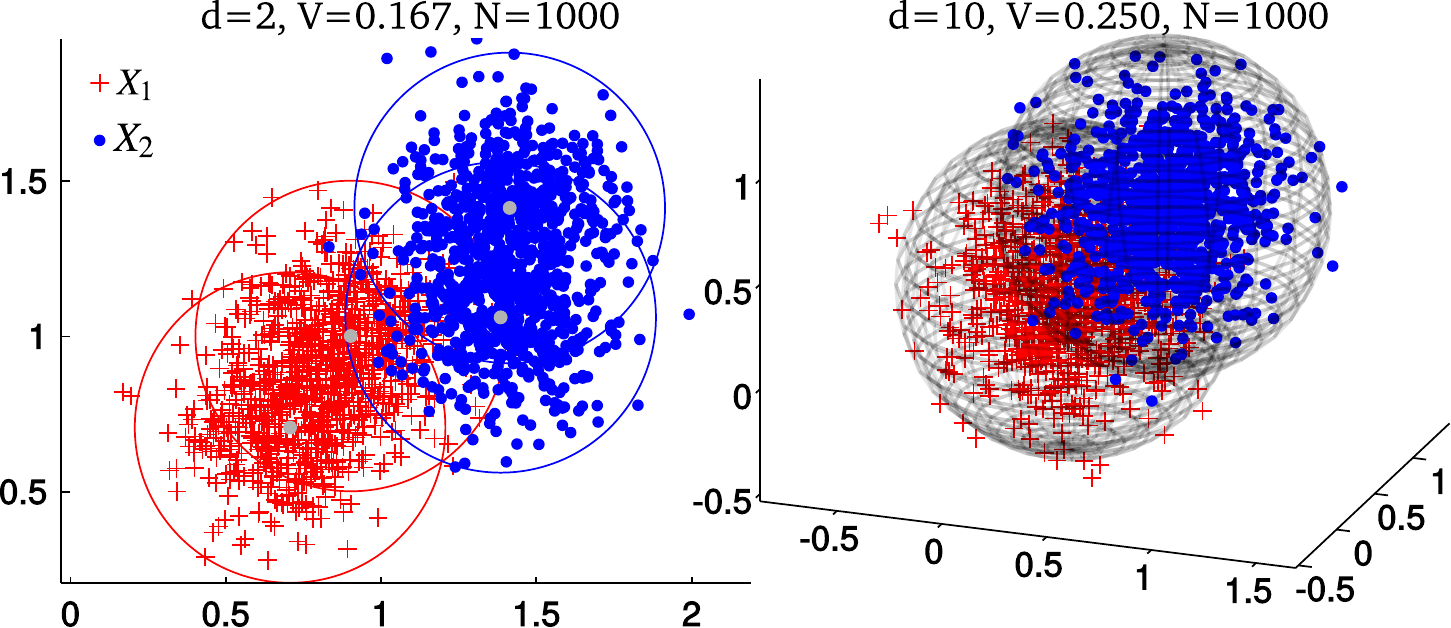}
  \caption{Examples of synthetic datasets generated. For the plot on the right only the first three dimensions are shown.}
\label{fig:syntheticdatasets}
 \end{figure}

\subsection{Experimental setup}


 
%

A stratified $10$-fold technique has been performed to divide all datasets. Each experiment is repeated 3 times in order to obtain robust results (except for deterministic methods).
The results are taken as mean and standard deviation of the selected measures.
 The same seed is used for random number generation, meaning that the same patterns are removed from the dataset and created by over-sampling.  
The cost parameter of SVM-based methods was selected within the values $\{10^{-1}, 10^{0}, 10^1\}$ by means of a nested $3$-fold method with the training set. The kernel parameter has been cross-validated  within the values $\{10^{-1}, 10^{0}, 10^1\}$ for the SVM based methods. 
For all the methods using large-scale semi-supervised SVMs \cite{sindhwani2006large}, the regularisation parameters $w$ and $u$  were optimised within the values $\{10^{-1},10^0,10^1\}$ (also by means of a nested $3$-fold cross-validation). 
For easing the comparisons, the number of synthetically generated patterns is set to the same removed initially from the dataset. $k=5$ nearest neighbours were 
evaluated to generate synthetic samples. 
The Euclidean distance has been used for all the distance computations.

The parameter used for the over-sampling method in \cite{Perez2016} to control the dimensionality of the feature space has been cross-validated within the values \{0.25, 0.5, 0.75\}. 
The kernel width parameter associated to transductive methods (to construct the graph) has been set to the same value of the SVM kernel used. The rest of parameters have been set to default values.  

There are several minor modifications of these algorithms when using them for either small or imbalanced datasets.
 As stated before, in the case of imbalanced data, we introduce a new parameter for S$^3$VM methods, which controls the ratio of patterns assigned to the minority class. This class balancing parameter has been fixed to the initial class distribution (in the first and second experiments where the data is balanced) and cross-validated within the values \{0.5,0.7,0.9\} for the imbalanced datasets (where all the synthetically generated patterns are supposed to belong to the minority class, but where we need to allow a certain amount of errors, to fix label inconsistencies).
Moreover, for the case of graph-based algorithms, several issues have been noticed in imbalanced domains \cite{10.1109/ASONAM.2016.7752356}. To prevent this, we also use a class mass normalisation procedure to adjust the class distribution so that it matches the priors \cite{zhuGhahramLaff03semisup}.

%

  \begin{figure*}[ht!]
   \centering
     \includegraphics[width = 0.9\textwidth]{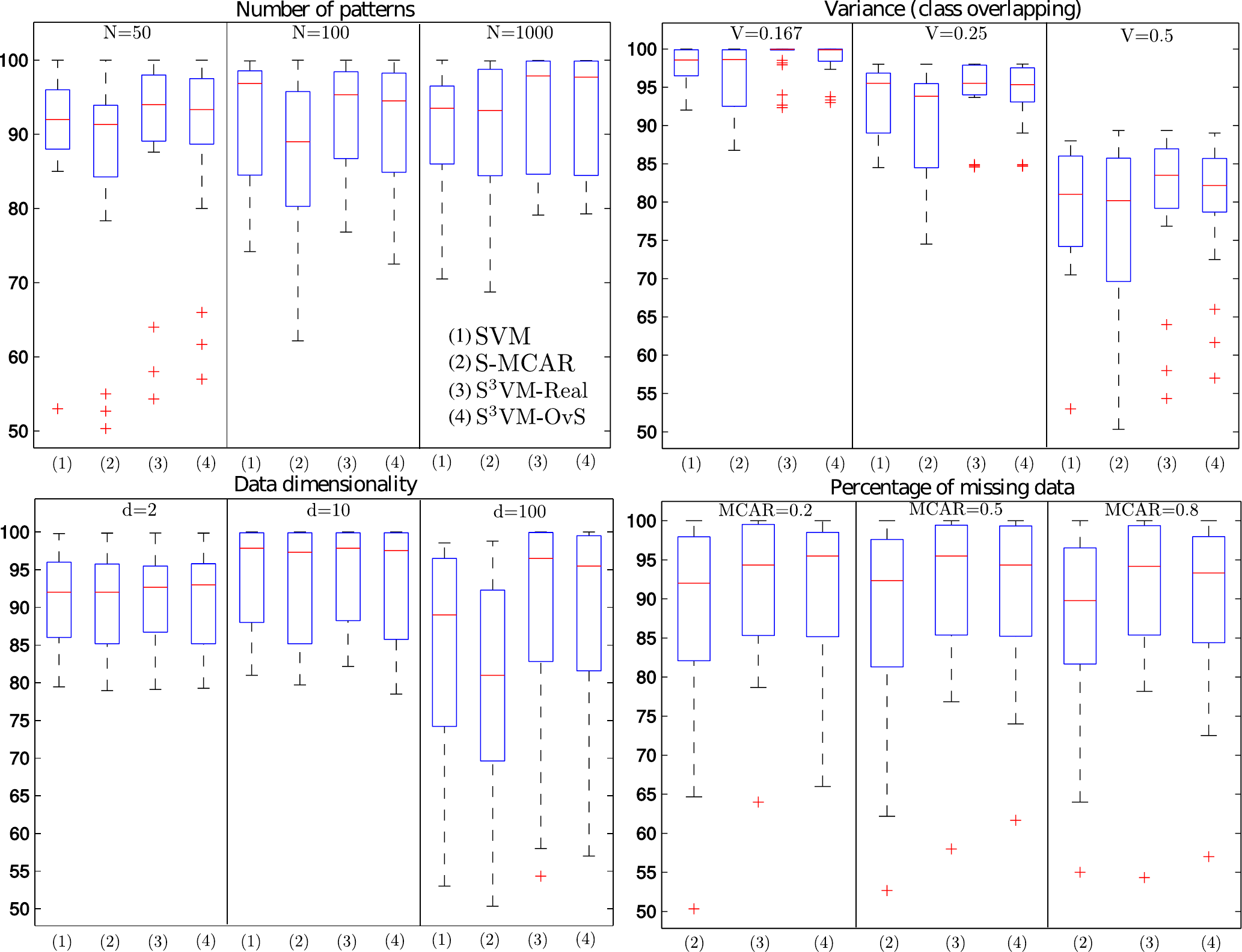}
  \caption{Box-plot of the mean test accuracy performance across different factors for the synthetic datasets (first experiment). }
\label{fig:allFactors}
 \end{figure*}

\subsection{Evaluation metrics}

The results have been reported in terms of two metrics: 
\begin{enumerate}
 \item Accuracy ($Acc$). However, given that for imbalanced cases this metric is not be the best option, we use the mean of the sensitivities per class (referred to as $MAcc$). 
\item The Geometric Mean of the sensitivities ($GM = \sqrt{S_\mathrm{p} \cdot S_\mathrm{n}}$) \cite{Kubat97addressingthe}, where $S_\mathrm{p}$ is the sensitivity for the positive class (ratio of correctly classified patterns considering only this class), and $S_\mathrm{n}$ is the sensitivity for the negative one.
\end{enumerate}
The measure for the parameter selection was $GM$ {given its robustness \cite{Kubat97addressingthe}. 

\subsection{Results}

Firstly, we test the influence of the number of patterns removed at random. \figurename{ \ref{fig:mcar}} shows the mean degradation in test performance for S-MCAR when changing the number of patterns removed from the benchmark datasets. As can be seen, all metrics experience a relatively large degradation.

 \subsubsection{First experiment: Synthetically generated datasets}

 $27$ synthetic datasets generated with \cite{Sanchez2013} are used. All of these datasets represent binary and perfectly balanced classification tasks, in which the data has been generated using a Normal distribution changing different parameters: 1) dimensionality of the input space (d, which is set to 2, 10 and 100 dimensions), 2) the number of patterns (N, set to 50, 100 and 1000) and 3) the variance of the data (V, controlling the overlapping between the classes and set to 0.167, 0.25 and 0.5). All combinations of these parameters have been explored. All the classes have been designed to be bi-modal. \figurename{ \ref{fig:syntheticdatasets}} shows two examples of the synthetic datasets generated. We test three ratios of patterns removed at random (MCAR): $0.2$, $0.5$ and $0.8$.

 For this experiment, we use four approaches: SVM (with the original dataset), S-MCAR (MCAR, no over-sampling), S$^3$VM with real unlabelled data (S$^3$VM-Real, for which the data that we remove is included again as unlabelled in the model) and our proposal using class-dependent over-sampling (S$^3$VM-OvS). Note that the comparison against SVM and S$^3$VM-Real is only for comparison purposes and not strictly fair, since the classifier has access to all the real data, which is not the case for S-MCAR and S$^3$VM-OvS.

   \begin{table*}[!]
 \begin{tiny}
\setlength{\tabcolsep}{7pt}
 \caption{Mean $Acc$ performance for the synthetic datasets considered.}
\begin{center}
   \begin{tabular}{  cccc cccc }
   \hline
   \hline
 d	&	N	&	V	&	MCAR	&	SVM	&	S-MCAR	&	S$^3$VM-Real	&	S$^3$VM-OvS	\\ \hline
2	&	50	&	0.167	&	0.2	&$	92.00	$&$	92.00	$&$	\mathit{92.33}	$&$	\mathbf{93.76}	$\\
2	&	50	&	0.167	&	0.5	&$	92.00	$&$	93.33	$&$	\mathbf{94.00}	$&$	93.33	$\\
2	&	50	&	0.167	&	0.8	&$	92.00	$&$	91.33	$&$	92.67	$&$	\mathbf{93.00}	$\\
2	&	50	&	0.25	&	0.2	&$	\mathbf{96.00}	$&$	94.00	$&$	94.00	$&$	92.48	$\\
2	&	50	&	0.25	&	0.5	&$	\mathbf{96.00}	$&$	93.67	$&$	93.67	$&$	94.33	$\\
2	&	50	&	0.25	&	0.8	&$	\mathbf{96.00}	$&$	93.00	$&$	94.00	$&$	93.33	$\\
2	&	50	&	0.5	&	0.2	&$	88.00	$&$	\mathbf{89.33}	$&$	\mathbf{89.33}	$&$	88.67	$\\
2	&	50	&	0.5	&	0.5	&$	88.00	$&$	87.67	$&$	87.59	$&$	\mathbf{88.33}	$\\
2	&	50	&	0.5	&	0.8	&$	88.00	$&$	88.67	$&$	\mathbf{88.96}	$&$	88.33	$\\
2	&	100	&	0.167	&	0.2	&$	\mathbf{99.80}	$&$	98.33	$&$	98.50	$&$	98.33	$\\
2	&	100	&	0.167	&	0.5	&$	\mathbf{99.80}	$&$	98.00	$&$	98.17	$&$	97.83	$\\
2	&	100	&	0.167	&	0.8	&$	\mathbf{99.80}	$&$	97.00	$&$	97.93	$&$	98.00	$\\
2	&	100	&	0.25	&	0.2	&$	84.50	$&$	\mathbf{95.83}	$&$	95.33	$&$	95.67	$\\
2	&	100	&	0.25	&	0.5	&$	84.50	$&$	95.50	$&$	95.50	$&$	\mathbf{95.83}	$\\
2	&	100	&	0.25	&	0.8	&$	84.50	$&$	95.00	$&$	\mathbf{95.50}	$&$	95.33	$\\
2	&	100	&	0.5	&	0.2	&$	79.45	$&$	\mathbf{87.00}	$&$	\mathbf{87.00}	$&$	86.00	$\\
2	&	100	&	0.5	&	0.5	&$	79.45	$&$	86.33	$&$	\mathbf{86.83}	$&$	\mathbf{86.83}	$\\
2	&	100	&	0.5	&	0.8	&$	79.45	$&$	83.83	$&$	\mathbf{86.67}	$&$	84.50	$\\
2	&	1000	&	0.167	&	0.2	&$	97.50	$&$	99.87	$&$	\mathbf{99.88}	$&$	\mathbf{99.88}	$\\
2	&	1000	&	0.167	&	0.5	&$	97.50	$&$	\mathbf{99.87}	$&$	\mathbf{99.87}	$&$	99.85	$\\
2	&	1000	&	0.167	&	0.8	&$	97.50	$&$	99.84	$&$	\mathbf{99.87}	$&$	99.83	$\\
2	&	1000	&	0.25	&	0.2	&$	\mathbf{95.50}	$&$	84.80	$&$	84.77	$&$	84.90	$\\
2	&	1000	&	0.25	&	0.5	&$	\mathbf{95.50}	$&$	84.73	$&$	84.90	$&$	84.68	$\\
2	&	1000	&	0.25	&	0.8	&$	\mathbf{95.50}	$&$	84.40	$&$	84.58	$&$	84.72	$\\
2	&	1000	&	0.5	&	0.2	&$	\mathbf{86.00}	$&$	79.23	$&$	79.40	$&$	79.63	$\\
2	&	1000	&	0.5	&	0.5	&$	\mathbf{86.00}	$&$	79.15	$&$	79.12	$&$	79.40	$\\
2	&	1000	&	0.5	&	0.8	&$	\mathbf{86.00}	$&$	78.97	$&$	79.43	$&$	79.27	$\\\hline
10	&	50	&	0.167	&	0.2	&$	\mathbf{100.00}	$&$	\mathbf{100.00}	$&$	\mathbf{100.00}	$&$	\mathbf{100.00}	$\\ 
10	&	50	&	0.167	&	0.5	&$	\mathbf{100.00}	$&$	\mathbf{100.00}	$&$	\mathbf{100.00}	$&$	\mathbf{100.00}	$\\
10	&	50	&	0.167	&	0.8	&$	\mathbf{100.00}	$&$	\mathbf{100.00}	$&$	\mathbf{100.00}	$&$	\mathbf{100.00}	$\\
10	&	50	&	0.25	&	0.2	&$	\mathbf{98.00}	$&$	\mathbf{98.00}	$&$	\mathbf{98.00}	$&$	97.67	$\\
10	&	50	&	0.25	&	0.5	&$	\mathbf{98.00}	$&$	97.33	$&$	\mathbf{98.00}	$&$	97.56	$\\
10	&	50	&	0.25	&	0.8	&$	\mathbf{98.00}	$&$	95.33	$&$	97.33	$&$	97.33	$\\
10	&	50	&	0.5	&	0.2	&$	88.00	$&$	\mathbf{89.33}	$&$	88.00	$&$	88.67	$\\
10	&	50	&	0.5	&	0.5	&$	88.00	$&$	86.00	$&$	\mathbf{89.00}	$&$	\mathbf{89.00}	$\\
10	&	50	&	0.5	&	0.8	&$	88.00	$&$	83.67	$&$	\mathbf{89.33}	$&$	80.00	$\\
10	&	100	&	0.167	&	0.2	&$	99.90	$&$	\mathbf{100.00}	$&$	\mathbf{100.00}	$&$	\mathbf{100.00}	$\\
10	&	100	&	0.167	&	0.5	&$	99.90	$&$	\mathbf{100.00}	$&$	\mathbf{100.00}	$&$	\mathbf{100.00}	$\\
10	&	100	&	0.167	&	0.8	&$	99.90	$&$	\mathbf{100.00}	$&$	\mathbf{100.00}	$&$	\mathbf{100.00}	$\\
10	&	100	&	0.25	&	0.2	&$	\mathbf{97.85}	$&$	94.17	$&$	94.00	$&$	94.17	$\\
10	&	100	&	0.25	&	0.5	&$	\mathbf{97.85}	$&$	94.50	$&$	94.00	$&$	94.00	$\\
10	&	100	&	0.25	&	0.8	&$	\mathbf{97.85}	$&$	93.83	$&$	93.83	$&$	93.33	$\\
10	&	100	&	0.5	&	0.2	&$	\mathbf{84.70}	$&$	80.67	$&$	82.67	$&$	82.17	$\\
10	&	100	&	0.5	&	0.5	&$	\mathbf{84.70}	$&$	80.17	$&$	84.00	$&$	82.17	$\\
10	&	100	&	0.5	&	0.8	&$	\mathbf{84.70}	$&$	79.70	$&$	82.17	$&$	78.50	$\\
10	&	1000	&	0.167	&	0.2	&$	\mathbf{100.00}	$&$	99.90	$&$	99.90	$&$	99.90	$\\
10	&	1000	&	0.167	&	0.5	&$	\mathbf{100.00}	$&$	99.90	$&$	99.90	$&$	99.90	$\\
10	&	1000	&	0.167	&	0.8	&$	\mathbf{100.00}	$&$	99.90	$&$	99.90	$&$	99.90	$\\
10	&	1000	&	0.25	&	0.2	&$	93.50	$&$	97.82	$&$	\mathbf{97.85}	$&$	97.83	$\\
10	&	1000	&	0.25	&	0.5	&$	93.50	$&$	97.72	$&$	\mathbf{97.90}	$&$	97.85	$\\
10	&	1000	&	0.25	&	0.8	&$	93.50	$&$	97.72	$&$	\mathbf{97.88}	$&$	97.83	$\\
10	&	1000	&	0.5	&	0.2	&$	81.00	$&$	\mathbf{84.87}	$&$	84.68	$&$	84.78	$\\
10	&	1000	&	0.5	&	0.5	&$	81.00	$&$	\mathbf{84.90}	$&$	\mathbf{84.90}	$&$	84.68	$\\
10	&	1000	&	0.5	&	0.8	&$	81.00	$&$	84.47	$&$	\mathbf{84.97}	$&$	84.37	$\\ \hline
100	&	50	&	0.167	&	0.2	&$	92.00	$&$	93.00	$&$	\mathbf{100.00}	$&$	98.58	$\\
100	&	50	&	0.167	&	0.5	&$	92.00	$&$	92.33	$&$	\mathbf{100.00}	$&$	\mathbf{100.00}	$\\
100	&	50	&	0.167	&	0.8	&$	92.00	$&$	89.11	$&$	\mathbf{100.00}	$&$	97.33	$\\
100	&	50	&	0.25	&	0.2	&$	85.00	$&$	78.33	$&$	94.33	$&$	\mathbf{96.19}	$\\
100	&	50	&	0.25	&	0.5	&$	85.00	$&$	79.00	$&$	\mathbf{96.67}	$&$	92.00	$\\
100	&	50	&	0.25	&	0.8	&$	85.00	$&$	81.00	$&$	\mathbf{94.33}	$&$	89.00	$\\
100	&	50	&	0.5	&	0.2	&$	53.00	$&$	50.33	$&$	64.00	$&$	\mathbf{66.00}	$\\
100	&	50	&	0.5	&	0.5	&$	53.00	$&$	52.67	$&$	58.00	$&$	\mathbf{61.67}	$\\
100	&	50	&	0.5	&	0.8	&$	53.00	$&$	55.00	$&$	54.33	$&$	\mathbf{57.00}	$\\
100	&	100	&	0.167	&	0.2	&$	98.55	$&$	86.75	$&$	\mathbf{100.00}	$&$	\mathbf{100.00}	$\\
100	&	100	&	0.167	&	0.5	&$	98.55	$&$	89.00	$&$	\mathbf{100.00}	$&$	99.83	$\\
100	&	100	&	0.167	&	0.8	&$	98.55	$&$	92.17	$&$	\mathbf{100.00}	$&$	\mathbf{100.00}	$\\
100	&	100	&	0.25	&	0.2	&$	\mathbf{96.85}	$&$	81.17	$&$	95.67	$&$	95.50	$\\
100	&	100	&	0.25	&	0.5	&$	\mathbf{96.85}	$&$	78.67	$&$	96.50	$&$	94.50	$\\
100	&	100	&	0.25	&	0.8	&$	\mathbf{96.85}	$&$	74.50	$&$	94.17	$&$	93.00	$\\
100	&	100	&	0.5	&	0.2	&$	74.20	$&$	64.67	$&$	\mathbf{78.67}	$&$	77.50	$\\
100	&	100	&	0.5	&	0.5	&$	74.20	$&$	62.17	$&$	\mathbf{76.83}	$&$	74.00	$\\
100	&	100	&	0.5	&	0.8	&$	74.20	$&$	64.00	$&$	\mathbf{78.17}	$&$	72.50	$\\
100	&	1000	&	0.167	&	0.2	&$	96.50	$&$	98.80	$&$	\mathbf{99.95}	$&$	99.93	$\\
100	&	1000	&	0.167	&	0.5	&$	96.50	$&$	98.62	$&$	\mathbf{99.95}	$&$	99.93	$\\
100	&	1000	&	0.167	&	0.8	&$	96.50	$&$	96.93	$&$	\mathbf{99.93}	$&$	\mathbf{99.93}	$\\
100	&	1000	&	0.25	&	0.2	&$	89.00	$&$	95.30	$&$	97.88	$&$	\mathbf{98.02}	$\\
100	&	1000	&	0.25	&	0.5	&$	89.00	$&$	93.20	$&$	\mathbf{97.93}	$&$	97.71	$\\
100	&	1000	&	0.25	&	0.8	&$	89.00	$&$	89.78	$&$	\mathbf{97.90}	$&$	97.42	$\\
100	&	1000	&	0.5	&	0.2	&$	70.50	$&$	73.95	$&$	83.15	$&$	\mathbf{83.40}	$\\
100	&	1000	&	0.5	&	0.5	&$	70.50	$&$	72.25	$&$	\mathbf{83.50}	$&$	83.09	$\\
100	&	1000	&	0.5	&	0.8	&$	70.50	$&$	68.75	$&$	\mathbf{82.72}	$&$	81.08	$\\
\hline \hline
   \end{tabular}
 \end{center}
 \label{table:resultadossynthetic}
 \end{tiny}
 \end{table*}

 From this experiment, we had results for $27$ datasets with different characteristics for three different MCAR levels and four methods (a total of 324 individual results). These results can be seen in \tablename{ \ref{table:resultadossynthetic}}. To analyse these properly, we summarised these results independently per factor in \figurename{ \ref{fig:allFactors}} using box-plots.  
Some conclusions can be drawn: 
Firstly, the overlapping of the classes (variance factor) is the main factor contributing to performance degradation. If the data does not overlap (small variance), a high performance can be achieved even if we remove data (compare method (1) to (2)).
The same is applicable when data dimensionality is low, e.g. for d=2 and d=10 removing data is not problematic (again, compare method (1) to (2)).
 However, an important degradation is seen when d=100.
The removal of data especially affects small datasets (N=50 and N=100) but not when N=1000.
Concerning the proposed approach (S$^3$VM-OvS), similar results can be achieved using real unlabelled data (S$^3$VM-Real), which is a positive outcome. Both results are also close to the performance using the complete dataset (compare approaches (3) and (4) to (1)), which means that over-sampled data can replace real one, even when real data is labelled.
 In some cases, such as in high-dimensional datasets, the performance even surpasses the one obtained by the original data. The proposal not only helps with small datasets, but also with relatively large ones (N=1000), perhaps because in this scenario the amount of data helps simplify the over-sampling task by exploiting better the local information. Thus, we can conclude that the proposed methodology helps specially for high dimensional datasets independently of their size and class overlapping, and that its performance is stable with respect to the percentage of data that we removed (last factor).

    \begin{table}[ht!]
 \begin{scriptsize}
\setlength{\tabcolsep}{4pt}
 \caption{Mean ranking results for all the methods considered in the small sample size experiment (second experiment).}
\centering
   \begin{tabular}{  ccccccc }
   \hline
   \hline
  & \multicolumn{2}{c}{MCAR ($0.2$)}  & \multicolumn{2}{c}{MCAR ($0.5$)} & \multicolumn{2}{c}{MCAR ($0.8$)} \\ 
Ranking	 & $MAcc$	&	$GM$ &	$MAcc$	&	$GM$ &	$MAcc$	&	$GM$	\\ \hline
SVM	&$4.62 $ & $4.46$&$	4.12	$&$	{4.12}	$ & $\mathbf{2.96}$ & $ {3.31}$\\
S-MCAR	& $8.00$ & $8.04$&$	6.81	$&$	{6.77}	$ & ${6.81}$ & $ {6.62}$\\
SVM+OvS	& $ 6.27$  & $6.62$& $	{5.85}	$&$	{5.92}	$ & ${5.85}$ & ${5.69}$ \\ 
SVM+kOvS & $7.06$  & $7.02$	&$	{7.19}	$&$	{7.08}	$ & ${6.54}$ & ${6.73}$ \\ \hline
\multicolumn{7}{c}{Transductive graph-based approaches}\\ \hline
Real unlab. data	& $9.52 $&$ 9.61$&$	10.04	$&$	10.00	$ & $9.98$ & $ 9.94$\\
Class dep. OvS	& $ 8.67$ &$8.85 $ &$	{9.27}	$&$	{9.35}$	& ${9.60}$ & ${9.60}$ \\ 
Class indep. OvS & $8.37$ & $8.35$	&$	{9.35}	$&$	{9.69}$	& ${9.69}$ & ${ 9.88}$ \\ \hline
\multicolumn{7}{c}{S$^3$VM approaches (proposed)}\\ \hline
Real unlab. data & $3.47 $&$3.35 $&$	\mathit{3.15}	$&$	\mathit{3.15}	$ & $\mathit{3.23}$ & $ \mathbf{3.04}$\\
Class dep. OvS & $\mathit{3.21}$	& $\mathit{3.27}$&	${3.54}	$&$	{3.42}$	& ${4.19}$ & ${4.27}$ \\ 
Class indep. OvS & $3.58$ & $3.38$	&$	{4.00}	$&$	{3.81}$	& ${3.88}$ & ${3.77}$ \\ 
Ensemble	& $\mathbf{3.15} $&$\mathbf{3.06} $&$	\mathbf{2.69}	$&$	\mathbf{2.69}$	& ${3.27}$ & $\mathit{ 3.15}$ \\ 
 \hline
\multicolumn{7}{c}{Friedman's test} \\  \hline
\multicolumn{7}{c}{Confidence interval $C_0 = (0, F_{(\alpha=0.05)} = 1.87)$} \\
\multicolumn{7}{c}{F-value MCAR =$0.2$: ${MAcc}$: $31.06 \notin C_0$, ${GM}$: $35.59\notin C_0$} \\
\multicolumn{7}{c}{F-value MCAR =$0.5$: ${MAcc}$: $49.21 \notin C_0$, ${GM}$: $57.40\notin C_0$} \\
\multicolumn{7}{c}{F-value MCAR =$0.8$: ${MAcc}$: $54.66 \notin C_0$, ${GM}$: $10.03\notin C_0$} \\
\hline \hline
   \end{tabular}
 \label{table:resultados1test}
 \end{scriptsize}
 \end{table}

  \figurename{ \ref{fig:N}} shows the effect of changing the number of synthetically generated patterns for S$^3$VM-OvS. It can be seen that similar results are obtained independently of the number of generated patterns. The percentage of data missing at random has, however, an important impact.
  
   \begin{figure}[ht!]
   \centering
     \includegraphics[width = 0.4\textwidth]{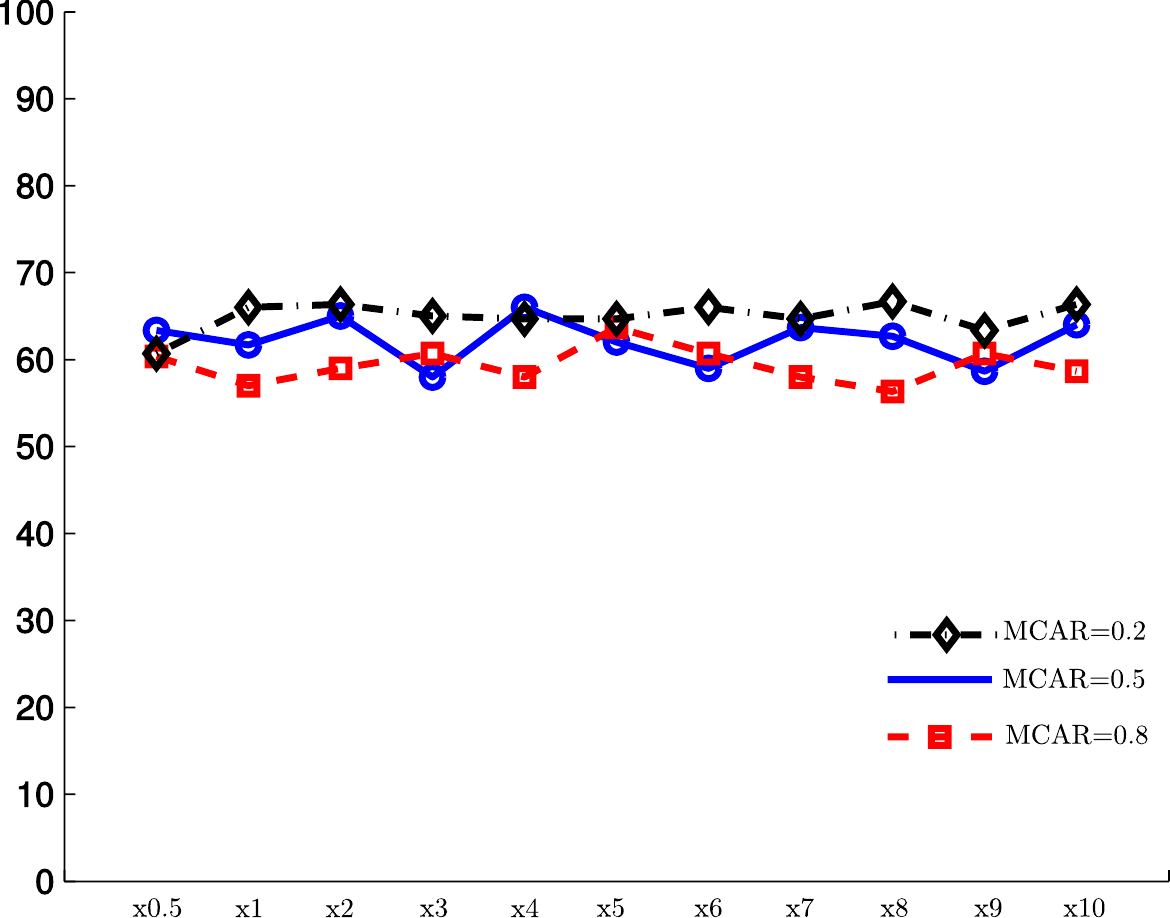}
  \caption{Mean $Acc$ performance for the dataset with parameters d=100, n=50 and V=0.5 using the S$^3$VM-OvS method changing the number of synthetically generated data (x0.5, x1, x2, \ldots, x10).  }
\label{fig:N}
 \end{figure}


\subsubsection{Second experiment: Small sample size}
For this experiment, we artificially reduce the size of the benchmark datasets (again testing a proportion of $0.2$, $0.5$ and $0.8$ reduction). Because of the amount of results we only provide the test mean ranking (the lower the better) in \tablename{ \ref{table:resultados1test}}. It can be seen that the
test rejects the null-hypothesis that all of the algorithms perform similarly in mean ranking
for all cases.
As mentioned before, here, we also include two over-sampling approaches from the literature: SVM+OvS \cite{Chawla02smote:synthetic} and SVM+kOvS \cite{Perez2016} and test transductive approaches to label synthetic data. Again, we compare several strategies: class-dependent and independent over-sampling, the introduction of real unlabelled data in the S$^3$VM model for comparison purposes and an ensemble. Note that both SVM and methods based on real unlab. data are unrealistic and only used as a baseline. 
Several conclusions can be drawn: Comparing all over-sampling approaches and S-MCAR it can be seen that a convex combination of patterns can be successfully used to generate synthetic data. 
 The use of part of the real data as unlabelled also improves the result to a reasonable extent: it is better than standard data over-sampling and if the number of data is not extremely low even better than use the original dataset, which may indicate that there might be some noise in the labels.
 The combination of over-sampling and semi-supervised learning approaches is promising and can be applied within each class or using all data independently of their labels, reaching in most cases the baseline performance of the use of the entire dataset. 
    Observing individual results we noticed that for the smallest datasets it is better to use all patterns for over-sampling, while for bigger datasets the best approach is to do over-sampling dependent on the class. 
  In general, transductive graph-based approaches do not report acceptable results, maybe because they highly depend on the design of a graph or because these techniques precise a larger amount of data.  
  Finally, the introduction of diversity in an ensemble by the use of a stochastic convex combination of patterns is very promising, improving in most cases the results achieved with the original complete dataset.

   \begin{table}[h]
 \begin{scriptsize}
\setlength{\tabcolsep}{10pt}
 \caption{Mean test ranking results for all the methods considered in the imbalanced experiment (third experiment).}
\centering
   \begin{tabular}{  ccccc }
   \hline
   \hline
   & \multicolumn{2}{c}{MCAR ($0.5$)} & \multicolumn{2}{c}{MCAR ($0.8$)} \\
Ranking	& 		$MAcc$	&	$GM$ &	$MAcc$	&	$GM$	\\ \hline
SVM	& $	\mathit{3.27}	$&$	\mathit{3.17}	$ & $\mathbf{2.50}$ & $ \mathit{2.65}$\\
S-MCAR	&$	{6.50}	$&$	{6.69}	$ & ${6.92}$ & ${6.96}$ \\
SVM+OvS	&$	4.08	$&$	4.08	$ & $4.56$ & $ 4.46$\\
SVM+kOvS	&$	{3.88}	$&$	{3.61}$	& ${3.54}$ & ${ 3.62}$ \\ \hline
\multicolumn{5}{c}{Transductive graph-based approaches}\\ \hline
Class dep. OvS	&$	{4.00}	$&$	{4.02}	$ & ${4.15}$ & ${4.27}$ \\ \hline
\multicolumn{5}{c}{Proposed S$^3$VM approaches}\\ \hline
Class dep. OvS	&$	\mathbf{2.69}	$&$	\mathbf{2.88}	$ & $\mathit{2.67}$ & $ \mathbf{2.38}$\\ 
Class indep. OvS	&$	{3.58}	$&$	{3.54}	$ & $3.67$ & $ 3.65$\\ \hline
\multicolumn{5}{c}{Friedman's test} \\  \hline
\multicolumn{5}{c}{Confidence interval $C_0 = (0, F_{(\alpha=0.05)} = 2.16)$} \\
\multicolumn{5}{c}{F-value MCAR=$0.5$: ${MAcc}$: $11.25 \notin C_0$, ${GM}$: $12.93\notin C_0$} \\
\multicolumn{5}{c}{F-value MCAR=$0.8$: ${MAcc}$: $20.49 \notin C_0$, ${GM}$: $24.10\notin C_0$} \\
\hline \hline
   \end{tabular}
 \label{table:resultados3test}
 \end{scriptsize}
 \end{table}
%

%
%
%
 
To quantify whether a statistical difference exists among the
algorithms studied, a procedure is employed to compare multiple classifiers in multiple datasets \cite{Demsar2006}.  \tablename{ \ref{table:resultados1test}} shows the result of applying  the non-parametric statistical Friedman's test (for a significance level of $\alpha=0.05$) to the mean $Acc$ and $GM$ rankings. The test rejects the null-hypothesis that all algorithms perform similarly in mean ranking for both metrics (for $GM$ the differences are larger).
 This table shows that the best results in $Acc$ and $GM$ are obtained obviously by the SVM technique (which does not entail any loss of information). This method is followed then by the semi-supervised S$^3$VM-MCAR+USP for $GM$ and SVM-MCAR+LSPFS for $Acc$. These results demonstrate that standard SMOTE (SVM-MCAR+LSP) creates label inconsistencies and that these can be avoided by the use of over-sampling in the feature space or a semi-supervised approach. Moreover, as the literature has suggested, graph-based (SVM-MCAR+LSPH) are not ideal for imbalanced datasets, although they still represent a better option than standard SMOTE. These results are accentuated when increasing the number of data missing at random.

\subsubsection{Third experiment: Imbalanced samples}

We also study the effect of our proposal in imbalanced classification setups. For this, we artificially induce this imbalance in our data by removing a percentage of patterns for the minority class. In this case, we test a subset of the methods that we used in the previous experiment (results shown in \tablename{ \ref{table:resultados3test}}). Again, we can see that SMOTE (SVM+OvS) can be improved, either by optimising the patterns to generate (SVM+kOvS) or the labels of the synthetic patterns (proposed approaches). It can also be seen that it is better to over-sample only the minority class (i.e. class dependent).

\section{Conclusions}

We explored the idea of introducing synthetic data as unsupervised information in semi-supervised support vector machines, where labels of synthetic data are treated as additional optimisation variables. Our experimental study has shown that: 1) synthetic patterns help when data is scarce with respect to the data dimensionality and can be used in a variety of cases as an alternative to collecting more data; 2) convex combination of input training data can be used for generating those synthetic samples, but these do not have to be necessarily labelled; and 3) the introduction of synthetic data as unsupervised knowledge can help to improve the classification in small, high-dimensional or imbalanced scenarios by acting as an inductive bias for the classifier.

Future work comprises testing such approach in a regression setting and with other semi-supervised learning approaches (e.g. the use of synthetic imaging data with autoencoders or deep belief networks). 

%

 \section{Acknowledgments}
This work was supported by the project TIN2017-85887-C2-1-P of the Spanish Ministry
of Economy and Competitiveness (MINECO) and FEDER funds (EU). 
PT was supported by the European Commission Horizon 2020 Innovative Training Network SUNDIAL (SUrvey Network for Deep Imaging Analysis and Learning), Project ID: 721463.

\fontsize{9.0pt}{10.0pt}
\selectfont

\bibliographystyle{aaai}
\bibliography{biblio,IEEEfull,references,mybibfile,biblio2,IEEEfull2}

%

\end{document}